\documentclass{spie}

\usepackage{cite}

\usepackage{times} % Times New Roman
\usepackage{indentfirst} % Indent first paragraphs

\usepackage[colorinlistoftodos,color=pink]{todonotes} % For fancy \todo{something}
\usepackage{xkeyval}
\presetkeys{todonotes}{inline}{}

\usepackage{graphicx}
\usepackage{amsmath}
\usepackage{amssymb}

\usepackage{multirow}
\usepackage{footnote}

\usepackage{wrapfig}

\usepackage{caption}
\title{Learning to Validate the Quality of Detected Landmarks}

\author{Wolfgang Fuhl, Enkelejda Kasneci 	
  \skiplinehalf
  \normalsize 
  Eberhard Karls Universit\"at T\"ubingen,
  72076 T\"ubingen, Germany \\
}

\begin{document}

\maketitle

\begin{abstract}
We present a new loss function for the validation of image landmarks detected via Convolutional Neural Networks (CNN). The network learns to estimate how accurate its landmark estimation is. This loss function is applicable to all regression-based location estimations and allows the exclusion of unreliable landmarks from further processing. In addition, we formulate a novel batch balancing approach which weights the importance of samples based on their produced loss. This is done by computing a probability distribution mapping on an interval from which samples can be selected using a uniform random selection scheme. We conducted experiments on the 300W, AFLW, and WFLW facial landmark datasets. In the first experiments, the influence of our batch balancing approach is evaluated by comparing it against uniform sampling. In addition, we evaluated the impact of the validation loss on the landmark accuracy based on uniform sampling. The last experiments evaluate the correlation of the validation signal with the landmark accuracy. All experiments were performed for all three datasets.
  \keywords{landmark detection, landmark validation, deep learning, batch creation}
\end{abstract}

\section{Introduction}
Landmark localization describes the process of determining the location of characteristic points of an object in an image. These points are usually characterized by a spatial relationship to each other such as the nose and eyes of a face. Under pose changes of the object, landmarks cannot change positions randomly but according to their relative positions towards each other. The problem of landmark localization is common in 2D face alignment, face reconstruction or gesture recognition~\cite{bulat2017far,koppen2018gaussian}. These are also preparatory steps for head pose estimation~\cite{zhu2012face,wu2008two}, emotion
estimation~\cite{walecki2016copula,li2017reliable} and face recognition~\cite{yang2017neural,liu2017sphereface}. The current state of the art in this area addresses the problems of accurate and robust landmark detection in real and constrained scenarios. Deep neural networks~\cite{huang2017densely,zhu2017unpaired} are currently in the focus of attention since they outperformed any other machine learning approach for common computer vision tasks. This includes landmark detection~\cite{bulat2017far,yang2017stacked} with a multitude of different network architectures, including convolution neuronal networks~\cite{sun2013deep}, auto encoders~\cite{zhang2014coarse}, recurrent networks~\cite{trigeorgis2016mnemonic}, residual networks~\cite{wingloss2018cvpr,dong2018style} and hourglass networks~\cite{bulat2017far}. Recent developments emphasize the importance of the loss function as a key element for improving the accuracy and robustness of landmark localization, such as the newly proposed Wing loss~\cite{wingloss2018cvpr}. Another novel architecture, a local to global context network proposed in~\cite{merget2018robust}, outputs heat map candidates and is therefore capable of detecting landmarks for multiple persons in an image. In this work, we address the challenge of facial landmark validation and batch balancing for training. The main contributions of this work are \textbf{1:} A novel loss function that trains the network directly to estimate the accuracy alongside with landmark location. The idea behind our validation loss is that we want the network to guess the reliability in terms of distance to the ground truth position.  \textbf{2:} An online data augmentation strategy that ranks training samples by their produced loss. This addresses the problem of pose-based data balancing~\cite{wingloss2018cvpr} by not only normalizing for the occurrences of head poses but also for (a priori unknown) challenges in the dataset like occlusion and illumination.

\section{Related work}
\label{sec:relwork}
State of the art methods employing deep learning for facial landmark detection is based on regression, i.e. the network estimates the correct landmark positions directly. We categorize those approaches by their employed architecture, data balancing procedure, cascading, and loss functions. \textbf{Architecture:} A straightforward way to use a CNN for direct 2D landmark regression was proposed in~\cite{wingloss2018cvpr,sun2013deep}. The input to such a network is a face image which is extracted based on preliminary applied face detection. Novel architectures such as  residual~\cite{wingloss2018cvpr} and recurrent networks~\cite{trigeorgis2016mnemonic} have already improved the state of the art in regression-based landmark detection. Hourglass networks~\cite{bulat2017far} have also led to an increase in both accuracy and robustness of landmark detection. However, instead of a position vector, its output is a heat map, where each pixel represents the probability to be a landmark position. \textbf{Data balancing:} Facial landmarks are subject to extreme pose variations. Head poses present in current datasets are however heavily biased towards frontal images; extreme orientations are rarely represented in the data. To overcome this limitation, multiview models were proposed, splitting the problem into frontal and profile faces. Those where already used in  traditional approaches, such as ASM~\cite{cootes1995active} and AAM~\cite{cootes2001active}, as well as for cascaded regression-based approaches~\cite{cao2014face,feng2015random}. In \cite{dong2018style}, a cycle GAN was used to generate images to handle style variations.

\section{Batch balancing}
\label{sec:batchbalance}
\begin{figure}[ht]
	\begin{center}
		\includegraphics[width=0.1\linewidth]{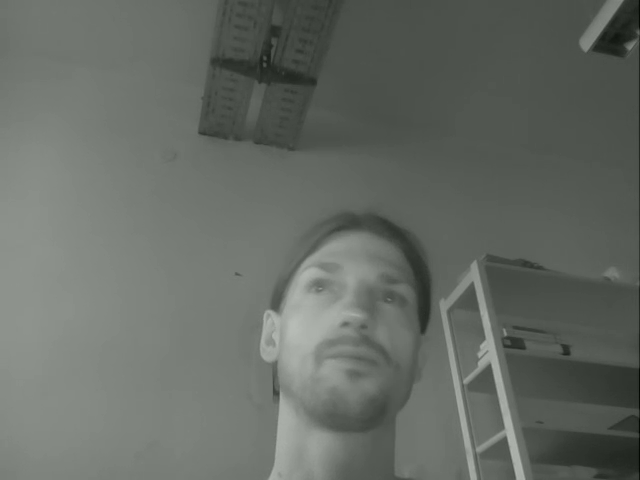}
		\includegraphics[width=0.1\linewidth]{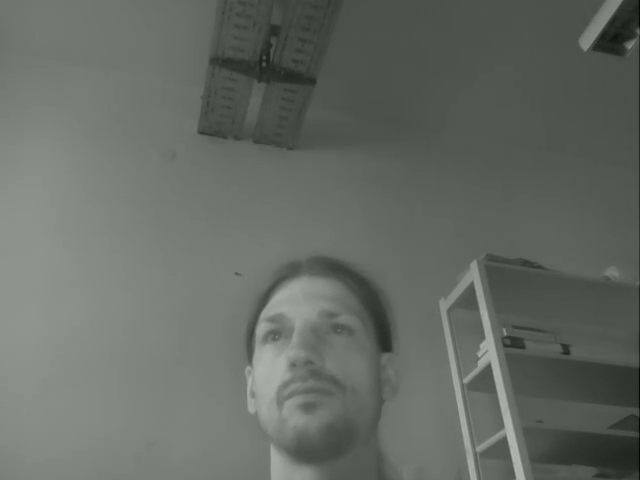}
		\includegraphics[width=0.1\linewidth]{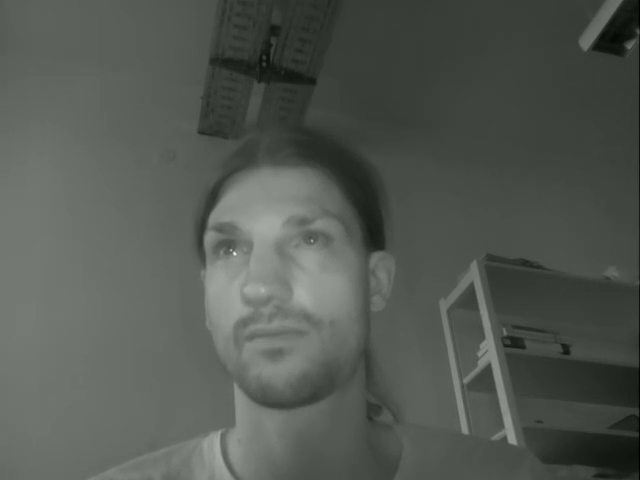}
		\includegraphics[width=0.1\linewidth]{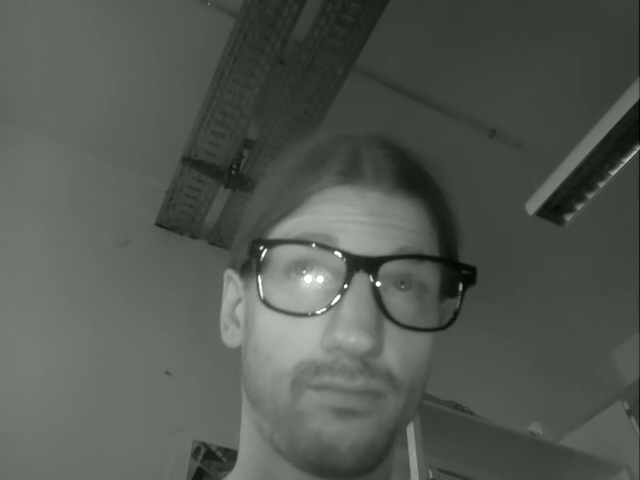}
		\includegraphics[width=0.1\linewidth]{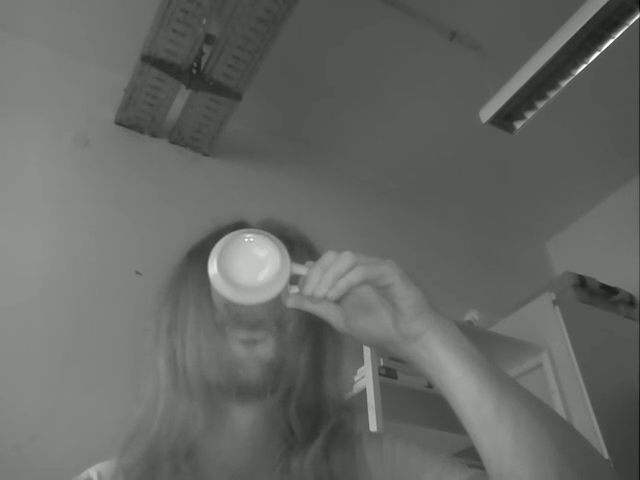}
		\includegraphics[width=0.1\linewidth]{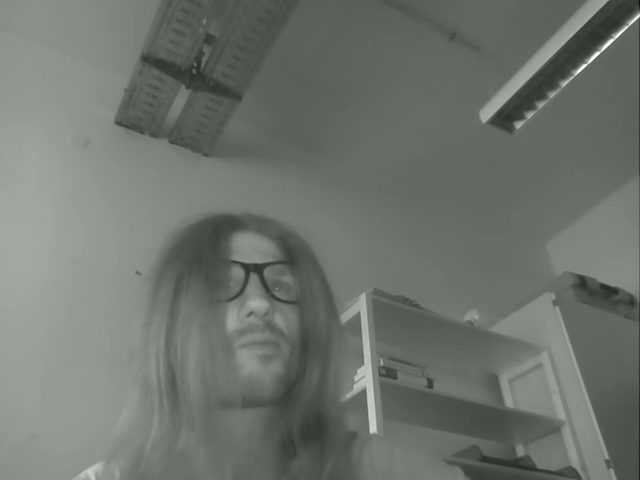}
		\includegraphics[width=0.1\linewidth]{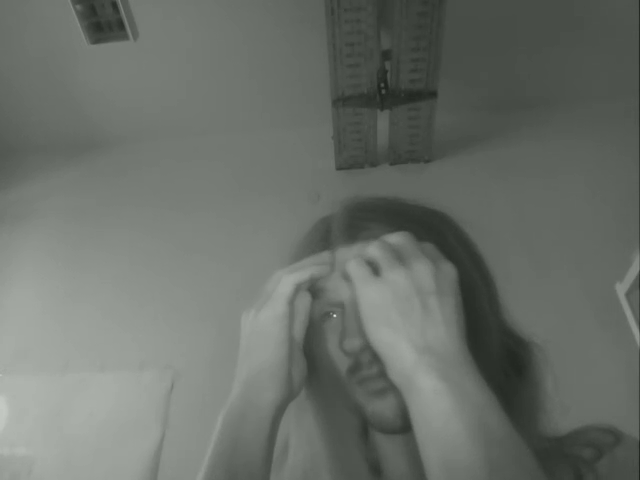}
		\includegraphics[width=0.1\linewidth]{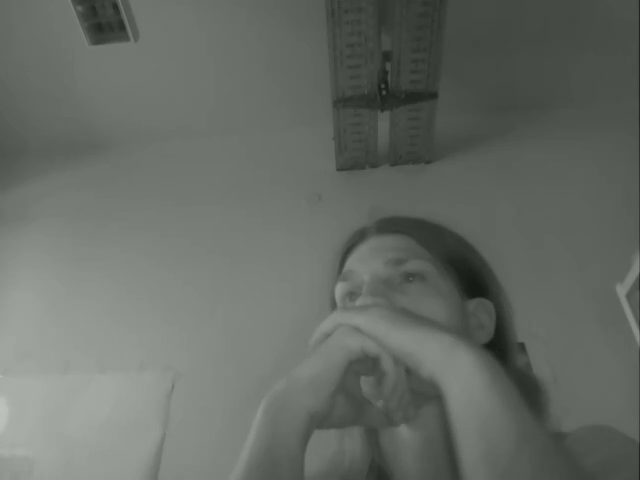}
	\end{center}
	\caption{Exemplary images which represent a balanced set based on the head pose.}
	\label{fig:batchbalance}
\end{figure}
Existing pose-based data balancing approaches~\cite{wingloss2018cvpr} are a valuable and effective procedure but have one major drawback: the data is balanced only based on the underrepresentation of some head poses with regard to a parameter of histogram bins.
Figure~\ref{fig:batchbalance} shows an example where data is already balanced based on the head pose. But additional challenges, such as reflections (top right) and three different types of occlusions are only represented by one sample each. Our idea is to use the produced loss of a training image to compute the probability of an image being included in the next training batch. Thereby, we can make the network learn especially the challenging cases, even if they are not annotated as such. 
\begin{figure}[ht]
	\begin{center}
		\includegraphics[width=0.4\linewidth]{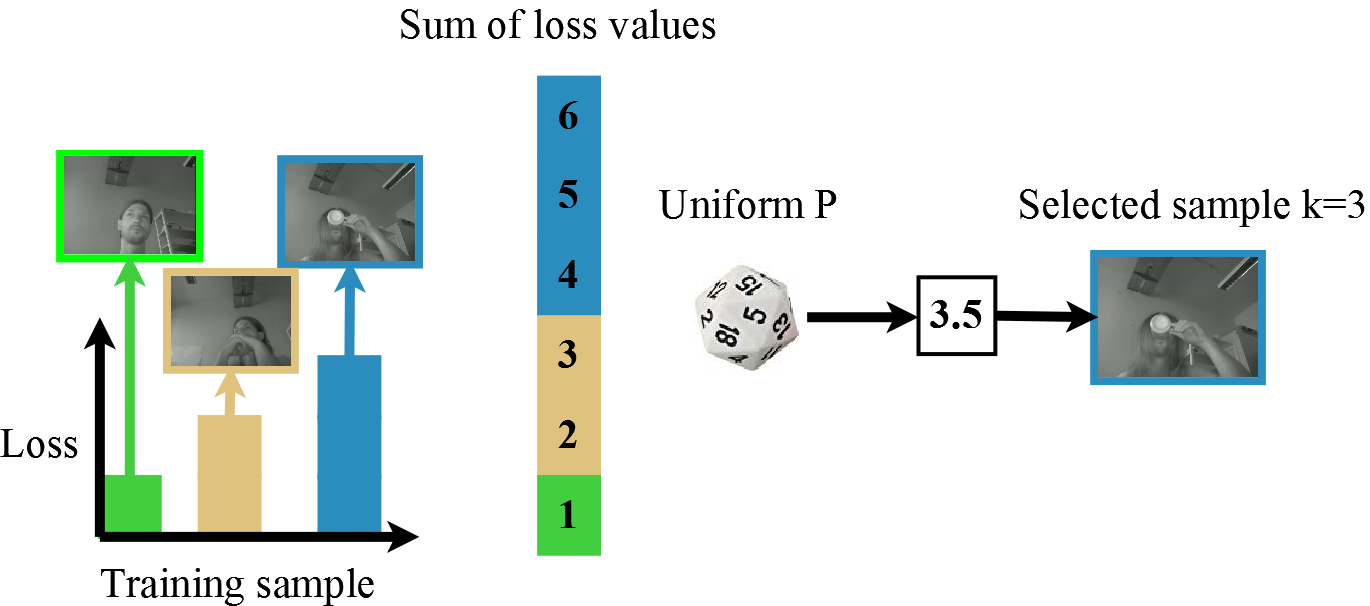}
	\end{center}
	\caption{The proposed ranking example where the produced loss represents the probability of a sample to be included in a training batch for the next epoch.}
	\label{fig:batchranking}
\end{figure}
Figure~\ref{fig:batchbalance} shows such a training set. Each sample, therefore, has an assigned value range which allows to create a new batch using a uniform random sampling.
\begin{equation}
B_{j}=S(k | \sum_{i=1}^{k}Loss_{i} \geq P([0;\sum_{i=1}^{n}Loss_{i}]))
\label{eq:PBBsampling}
\end{equation}
Equation~\ref{eq:PBBsampling} describes our \underline{L}oss \underline{P}robability based \underline{B}atch \underline{B}alancing (LPBB). For the selection of the $j$ training sample ($S$) in our batch ($B$) we first select a value in the range $[0;\sum_{i=1}^{n}Loss_{i}]$ using the uniform distribution $P$. $n$ is the amount of all samples in our training set. Afterward, we search for the $k$th sample which has a greater or equal sum of the loss values as our randomly selected value. For completeness, it should also be required that k is the smallest sum that satisfies this condition (Not integrated into the formula for simplification). This sample is than included in the batch. This also means that samples can be assigned twice in a batch but since we use online data augmentation, there is no limitation (Section~\ref{sec:trainandaugment}). Figure~\ref{fig:batchranking} shows the sample selection. First the sum is computed based on which the range for the uniform distribution is computed. Afterwards, the sample three is selected since it has a greater sum of loss values. For an exact determination of the probability based on the loss value, all data must be evaluated after each training batch. Since this represents a considerable effort, two different uses were considered in our evaluation (Section~\ref{sec:evaluation}). An exact description can be found in Section~\ref{sec:trainandaugment}.
\begin{figure}[ht]
	\begin{center}
		\includegraphics[width=0.8\linewidth]{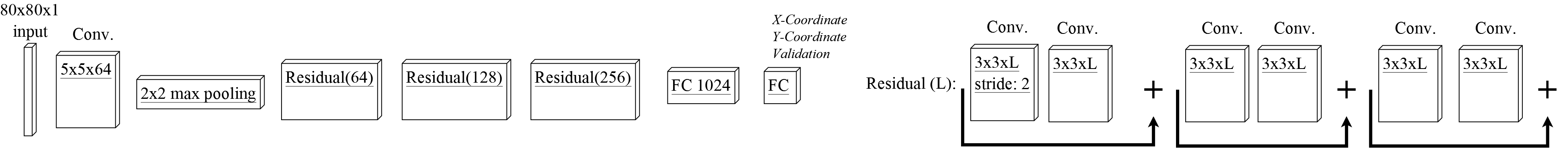}
	\end{center}
	\caption{The used network architecture for the proposed approach.}
	\label{fig:networkarch}
\end{figure}

\section{Validation loss formulation}
\label{sec:validloss}
The idea behind our validation loss is that we want the network to guess the expected distance between its prediction and the ground truth position as a proxy of the reliability of the prediction. This means that the network not only estimates the landmark position, it also evaluates its accuracy and provides this information in the additional output. Therefore, we compute the distance between the estimated and annotated position (L2, L1, or any task-specific distance metric) and use this as a training label for the next training iteration. For landmarks the convetional loss formulation is $\frac{1}{2} (gt_i-ev_i)^2$ for the $i$th index of the output neuron where $gt$ is the ground truth and $ev$ the estimated value. The derivative and therefore the gradient is $gt_i-ev_i$. Each landmark position is estimated using two neurons from the last layer as shown in Equation~\ref{eq:perlandmark}.
\begin{equation}
\begin{split}
LM_{j,x}=N_i&=gt_i-ev_i \\
LM_{j,y}=N_{i+1}&=gt_{i+1}-ev_{i+1}
\end{split}
\label{eq:perlandmark}
\end{equation}
$LM_{j,x}$ in Equation~\ref{eq:perlandmark} is the gradient for the x coordinate of the $j$th landmark and $LM_{j,y}$ the gradient for the y coordinate which are represented by neuron $N_i$ and $N_{i+1}$ respectively. We extended this formulation by a third value which is the validation signal. As metric between the ground truth position and the estimated value we used the Manhattan distance. The computation of the gradients for one landmark position using the validation loss is shown in Equation~\ref{eq:perlandmarkvloss}.
\begin{equation}
\begin{split}
N_i &=gt_i-ev_i \\
N_{i+1}&=gt_{i+1}-ev_{i+1}\\
N_{i+2}&=|gt_i-ev_i| + |gt_{i+1}-ev_{i+1}| - ev_{i+2}
\end{split}
\label{eq:perlandmarkvloss}
\end{equation}
As can be seen each landmark receives an additional neuron ($N_{i+2}$) which learns to evaluate the quality of the $x,y$ position estimation of the landmark. The higher this signal is, the less accurate is the landmark estimation. The advantages of this formulation are first than the CNN can learn to estimate the landmark quality end to end and that our formulation is not bound to a distance metric. The Manhattan distance ($|gt_i-ev_i| + |gt_{i+1}-ev_{i+1}|$) can also be replaced by the Euclidean distance ($\sqrt{(gt_i-ev_i)^2 + (gt_{i+1}-ev_{i+1})^2}$) or any other distance metric.

\section{System, training, and augmentation parameters}
\label{sec:trainandaugment}
For the training of our network, we used DLIB~\cite{king2009dlib}. The training was performed on a POWER8 server from IBM with four Nvidia Tesla P100 GPUs and 1TB of RAM. For the evaluation and measuring the inference runtime, we used a desktop computer with an NVIDIA GeForce GTX 1050ti, 16GB of RAM and an Intel i5-4570 CPU. The network architecture itself is shown in Figure~\ref{fig:networkarch} and consists of an initial convolution layer followed by a max pooling. Afterward, three residual blocks are used with 64, 128 and 256 layers, respectively. After each convolution, we used a batch normalization layer followed by a rectifier linear unit. The end of our architecture consists of two fully connected layers, wherein the last layer has $3*l$ outputs with the validation loss and $2*l$ without. $l$ represents the number of landmarks. This means that for the 300W~\cite{zhu2012face} dataset with 68 landmarks our output layer has $3*68=204$ neurons using the validation loss. These three values represent the x and y coordinate as well as the validation signal per landmark.

\textbf{Naming convention:} For our evaluation we trained four CNNs for each data set (300W~\cite{zhu2012face}, AFLW~\cite{koestinger11a}, and WFLW~\cite{wayne2018lab}). All CNNs follow the same architecture as shown in Figure~\ref{fig:networkarch}. NON: CNN using no validation signal and no batch balancing. NOBB: CNN using the validation signal and no batch balancing. BBO: CNN using the validation signal and batch balancing (PLBB) with an online update. BB: CNN using the validation signal and batch balancing (PLBB) with an update of the loss value for all training samples every 10\% of an epoch.

\textbf{Data augmentation:} For data augmentation, we used randomly added noise of up to 20\%. The face bounding box was randomly shifted by up to 20\% of the image in any direction (corners or between) and rotated between in the range [-1, 1] radian. Gaussian blur was added to the image in the range σ=[1.0, 1.3]. Additionally, we added occlusions which could cover up to 50\% of the image. The gray value of the occlusion was also randomly chosen in the range [0, 30] (dark), [200, 255] (bright) or randomly between [0, 255]. The last data augmentation was the contrast which was adjusted with a randomly selected value between [-30, 30].

\textbf{Training parameters:} Weight decay was set to $5\times10^{−4}$ and momentum to 0.9 with a batch size of 10. All images during training and evaluation were converted to grayscale. For the training of the NOBB, BBO, and BB networks we used the Manhattan distance for the validation loss as shown in Equation~\ref{eq:perlandmarkvloss}. All models were initialized with random values and trained in three steps. The first step used the L2 loss formulation for the x and y coordinate with a learning rate of $10^{-8}$. Each model was trained for $\approx2000$ epochs with this configuration. This was done since DLIB ends up in a loss of NaN for random initialization networks if the initial learning rate is set too high. Afterward, the learning rate was increased to $10^{-7}$ for additional $\approx2000$ epochs. For the second training step, we changed the loss formulation from L2 to L1 for the coordinate regression. The training was conducted for an additional $\approx2000$ epochs. Afterward, the learning rate was increased to $10^{-6}$ for all models with the exception of those trained on the WFLW~\cite{wayne2018lab} dataset. For the WFLW~\cite{wayne2018lab} dataset the learning rate was kept at $10^{-7}$. The training of all models was continued for an additional $\approx2000$ epochs also the models which are trained on WFLW. For the last and third part of our training, we changed the loss function to the wing loss~\cite{wingloss2018cvpr} for the coordinate regression. The wing loss function is a combination of a log loss and an L1 loss function. For large errors ($>w$) it behaves like the L1 loss function and for small errors ($<w$) like a log loss function. We used the parameters $\epsilon=2, w=10$ as in the original paper~\cite{wingloss2018cvpr} where $\epsilon$ is used to limit the curvature of the log function. The training was continued for an additional $900$ epochs where after each $300$ epochs the learning rate was decreased by $10^{-1}$ to a minimum of $10^{-8}$. For the models trained on the WFLW~\cite{wayne2018lab} dataset the learning rate was decreased to $10^{-8}$ after the first $500$ epochs and kept at this value for the remaining $400$ epochs.

\section{Evaluation}
\label{sec:evaluation}
\setlength{\tabcolsep}{0.4mm}
\begin{table}[h]
	\caption{Results in NME (smaller is better) for the networks NON and NOBB trained from scratch without batch balancing. Bold values are the best results.}
	\begin{center}
		\begin{tabular}{l|cc}
			\hline
			Dataset & NON & NOBB \\
			\hline\hline
			300W Full & \textbf{4.35} & 4.43 \\
			300W Common & \textbf{3.67} & 3.71 \\
			300W Challenging & \textbf{7.14} & 7.38 \\
			\hline
			AFLW Full & \textbf{1.74} & 1.76 \\
			\hline
			WFLW Full & \textbf{6.62} & 6.71 \\
			WFLW No challenge & \textbf{4.78} & 5.17 \\
			WFLW Blur & 6.99 & \textbf{6.94} \\
			WFLW Expression & \textbf{7.18} & 7.44 \\
			WFLW Illumination & \textbf{6.26} & 6.36 \\
			WFLW Makeup & 6.53 & \textbf{6.39} \\
			WFLW Occlusion & 7.49 & \textbf{7.20} \\
			WFLW Pose & 11.76 & \textbf{11.13} \\
			\hline
		\end{tabular}
	\end{center}
	\label{tbl:validimpacc}
\end{table}
In this section, we evaluate the proposed methodology and compare the resulting networks with the state of the art on the data sets 300W~\cite{zhu2012face}, AFLW~\cite{koestinger11a}, and WFLW~\cite{wayne2018lab}. First, the datasets are described. For the evaluation of our batch balancing approach and the comparison to the state of the art we used the 300W~\cite{zhu2012face}, AFLW~\cite{koestinger11a}, and WFLW~\cite{wayne2018lab} facial landmark dataset~\cite{zhu2012face}. All values are reported with a factor of $10^2$. The datasets are described in the following. \textbf{300W:} The dataset is an aggregation of multiple face datasets, namely LFPW~\cite{belhumeur2013localizing}, HELEN~\cite{le2012interactive}, AFW~\cite{zhu2012face} and XM2VTS~\cite{messer1999xm2vtsdb}. 68 landmarks are annotated semi automatically~\cite{sagonas2013semi}. The evaluation procedure is identical to~\cite{wingloss2018cvpr,dong2018style,ren2016face}. The training set consists of the annotated images from AFW and the training subset of LFPW and HELEN (3,148 images in total). For evaluation three test sets are considered. The common test set from LFPW and HELEN (554 images). A challenging test set consisting of 135 collected iBUG face images. All images together form the full test set (689 images). We report our results in normalized mean errors (NME). This corresponds to the pixel distance between detected and annotated landmark, normalized by the pixel distance between both eye centers (equal to the inter-pupillary distance if pupils are not annotated~\cite{wingloss2018cvpr,dong2018style,ren2016face}). \textbf{AFLW:} In our experiments, we used the AFLWFull protocol~\cite{zhu2016unconstrained}. The dataset consists of 20,000 training and 4,386 test images. Each image has 19 manually annotated landmarks and has been widely used for facial landmark localization benchmarking since it contains high pose variations(-90$^\circ$-90$^\circ$). We report our results in normalized mean errors (NME) which corresponds to the pixel distance between detected and annotated landmark, normalized by the width or height of the squared face bounding box. \textbf{WFLW:} The dataset consists of 7,500 images for training and 2,500 images for testing. Each image has 98 manually annotated facial landmarks. In addition, the test set is split up in different challenges e.g. occlusion, pose, make-up, illumination, blur, and expression. We followed the original protocol~\cite{wayne2018lab} and evaluated each challenge separately. The results are reported as normalized mean errors (NME). For the original protocol, the pixel distance between detected and annotated landmark is normalized using the inter-ocular distance.

\textbf{Impact of validation loss on the accuracy:} Table~\ref{tbl:validimpacc} shows the comparison of the models NON and NOBB. Both do not use the batch balancing but the model NOBB uses the proposed validation signal regression. As can be seen for the datasets 300W and AFLW the validation signal reduces the accuracy of the landmark regression. This due to the competition between the landmark position and its inaccuracy. For the dataset WFLW this is slightly different. Overall the model NON is more accurate but not for all challenges. For blur, makeup, occlusion, and pose the model NOBB outperforms NON.

\setlength{\tabcolsep}{0.4mm}
\begin{table}
	\caption{Average normalized mean error on the 300W dataset in comparison to the state of the art.}
	\begin{center}
		\begin{tabular}{l|ccc}
			\hline
			Method & Common & Challenging & Full \\
			\hline\hline
			TR-DRN~\cite{lv2017deep} & 4.36 & 7.56 & 4.99 \\
			LAB (8-stack)~\cite{lv2017deep} & 3.42 & 6.98 & 4.12 \\
			CNN-6(Wing)~\cite{wayne2018lab} & 3.35 & 7.20 & 4.04 \\
			CNN-6/7(Wing)~\cite{wayne2018lab} & 3.27 & 7.18 & 4.10 \\
			$SAN_{OD}$~\cite{dong2018style} & 3.41 & 7.55 & 4.24 \\
			$SAN_{GT}$~\cite{dong2018style} & 3.34 & 6.60 & 3.98 \\
			\hline
			NON & 3.67 & 7.14 & 4.35 \\
			NOBB & 3.71 & 7.38 & 4.43 \\
			BBO & 3.34 & 6.77 & 3.93 \\
			BB  & \textbf{3.01} & \textbf{6.51} & \textbf{3.69} \\
			\hline
		\end{tabular}
	\end{center}
	\label{tbl:cmp300w}
\end{table}

\setlength{\tabcolsep}{0.4mm}
\begin{table}[h]
	\caption{Average normalized mean error on the AFLW dataset in comparison to the state of the art.}
	\begin{center}
		\begin{tabular}{l|c}
			\hline
			Method & Full \\
			\hline\hline
			LAB w/o boundary~\cite{wayne2018lab} & 1.85\\
			CNN-6 (Wing)~\cite{wingloss2018cvpr} & 1.83 \\
			CNN-6/7 (Wing)~\cite{wingloss2018cvpr} & 1.65 \\
			$SAN_{GT}$~\cite{dong2018style} & 1.91 \\
			\hline
			NON & 1.74 \\
			NOBB & 1.76 \\
			BBO & 1.61 \\
			BB  & \textbf{1.56} \\
			\hline
		\end{tabular}
	\end{center}
	\label{tbl:cmpaflw}
\end{table}
\textbf{Impact of PLBB:} Table~\ref{tbl:cmp300w} shows the results on the 300W dataset. Comparing the results of NOBB to BBO and BB it can be seen that our batch balancing approach improves the results further. For BBO the results are slightly worse compared to BB because the loss distribution is less accurate. However, the computational effort for BBO is only one fifth. Comparing both (BB and BBO) to the state-of the art it can be seen that they are equally or more accurate. The models CNN-6~\cite{wayne2018lab} and CNN-6/7~\cite{wayne2018lab} where proposed together with the wing loss which we are using too. The authors additionally proposed a dataset balancing strategy (PDB) which shows to be effective by comparing these two networks to our NON and NOBB models which do not use any batch balancing strategy. The comparison of the CNN-6/7~\cite{wayne2018lab} to our BBO model it can be seen that the main improvement of our batch balancing strategy is in the challenging part of the 300W dataset. This is due to similar images in the training set produce higher loss values and are more likely to be in a batch. For the LAB~\cite{lv2017deep} landmark detection, there are also results using the ground truth boundary information on the 300W dataset. This combination together with the pretrained ResNet-50~\cite{wayne2018lab} is compared to our approach in Table~\ref{tbl:cmplargenets} where our model uses the validity signal to rule out inaccurate landmarks. Table~\ref{tbl:cmpaflw} shows the comparison of our approch on the AFLW dataset. As in Table~\ref{tbl:cmp300w} it can be seen that the batch balancing approach used for CNN-6/7~\cite{wingloss2018cvpr} is effective in comparison to our models NON and NOBB. The proposed batch balancing approach used in our models BBO and BB further improves the results on the AFLW dataset wherein compared to Table~\ref{tbl:cmp300w}, BBO is slightly more accurate than the network CNN-6/7~\cite{wingloss2018cvpr}. The results for LAB~\cite{wayne2018lab} with boundary information on the AFLW dataset will be shown in Table~\ref{tbl:cmplargenets} as well as the results for the pretrained ResNet-50~\cite{wingloss2018cvpr}.

\setlength{\tabcolsep}{0.4mm}
\begin{table}
	\caption{Average normalized mean error on the WFLW dataset in comparison to the state of the art.}
	\begin{center}
		\begin{tabular}{l|ccccccc}
			\hline
			& Full & Blur & Expression & Illumination & Makeup & Occlusion & Pose \\
			\hline\hline
			ESR~\cite{cao2014face}& 11.13 & 12.20 & 11.47 & 10.49 & 11.05 & 13.75 & 25.88 \\
			SDM~\cite{xiong2013supervised}& 10.29 & 11.28 & 11.45 & 9.32 & 9.38 & 13.03 & 24.10 \\
			CFSS~\cite{zhu2015face}& 9.07 & 9.96 & 10.09 & 8.30 & 8.74 & 11.76 & 21.36 \\
			DVLN~\cite{wu2017leveraging}& 6.08 & 6.88 & 6.78 & 5.73 & 5.98 & 7.33 & 11.54 \\
			LAB~\cite{wayne2018lab}& \textbf{5.27} & 6.32 & \textbf{5.51} & \textbf{5.23} & \textbf{5.15} & 6.79 & 10.24 \\
			\hline
			NON & 6.62 & 6.99 & 7.18 & 6.26 & 6.53 & 7.49 & 11.76\\
			NO & 6.71 & 6.94 & 7.44 & 6.36 & 6.39 & 7.20 & 11.13 \\
			BBO & 5.86 & 6.01 & 6.28 & 5.53 & 5.68 & 6.28 & 9.67 \\
			BB & 5.54 & \textbf{5.71} & 5.94 & 5.26 & 5.37 & \textbf{5.94} & \textbf{9.24}\\
			\hline
		\end{tabular}
	\end{center}
	\label{tbl:cmpwflw2}
\end{table}

\setlength{\tabcolsep}{0.4mm}
\begin{table}
	\caption{The results of our approach using different amounts of landmark positions based on the validation signal on the 300W dataset. For the model NON, we randomly selected 10\%, 20\% and 30\% as a baseline since it has no validation signal.}
	\begin{center}
		\begin{tabular}{cl|ccc}
			\hline
			Discard & Method & Common & Challenging & Full \\
			\hline\hline
			\multirow{4}{*}{10\%} & NON & 3.72 & 7.23 & 4.41 \\
			& NOBB  & 3.32 & 6.49 & 3.94 \\
			& BBO  & 2.91 & 6.02 & 3.52 \\
			& BB  & 2.72 & 5.79 & 3.32 \\
			\hline
			\multirow{4}{*}{20\%} & NON & 3.72 & 7.23 & 4.40 \\
			& NOBB  & 2.99 & 5.76 & 3.53 \\
			& BBO  & 2.64 & 5.40 & 3.18 \\
			& BB  & 2.48 & 5.18 & 3.01 \\
			\hline
			\multirow{4}{*}{30\%} & NON & 3.71 & 7.18 & 4.39 \\
			& NOBB  & 2.72 & 5.17 & 3.20 \\
			& BBO  & 2.42 & 4.87 & 2.90 \\
			& BB  & 2.27 & 4.66 & 2.74 \\
			\hline
		\end{tabular}
	\end{center}
	\label{tbl:cmpstoavl}
\end{table}

Table~\ref{tbl:cmpwflw2} shows the results of our models in comparison to the state of the art. It can be seen that overall the LAB~\cite{wayne2018lab} model performs best on the WFLW dataset with which it was published together. It has to be mentioned that the landmarks on this dataset were annotated so that they always lie on the boundary of the face. This makes the LAB~\cite{wayne2018lab} model especially effective on this dataset. In comparison to our models it can be seen that our batch balancing approach proof to be effective especially for the challenging pose dataset. Both models BB and BBO outperform the LAB~\cite{wayne2018lab} approach on this challenge. In addition, blur as well as occlusions show to be more accurate using our batch balancing. This has the same reason as for the challenging iBUG images from the 300W dataset. For all approaches in the Table~\ref{tbl:cmpwflw2} it can be seen that blur, occlusion and pose are the most inaccurate which means that similar images produce the highest loss values.

\begin{figure}[ht]
	\begin{center}
		\includegraphics[width=0.27\linewidth]{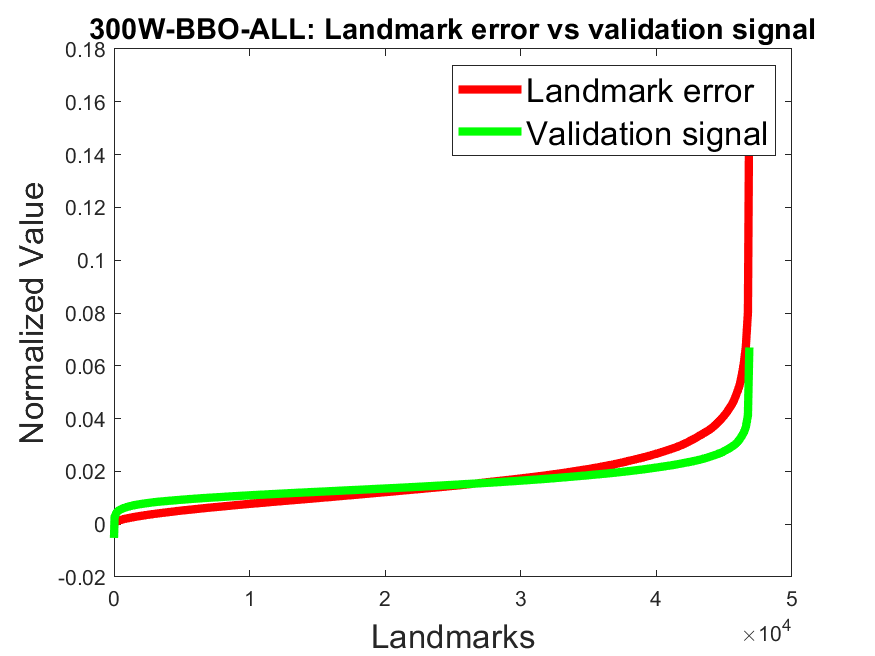}
		\includegraphics[width=0.27\linewidth]{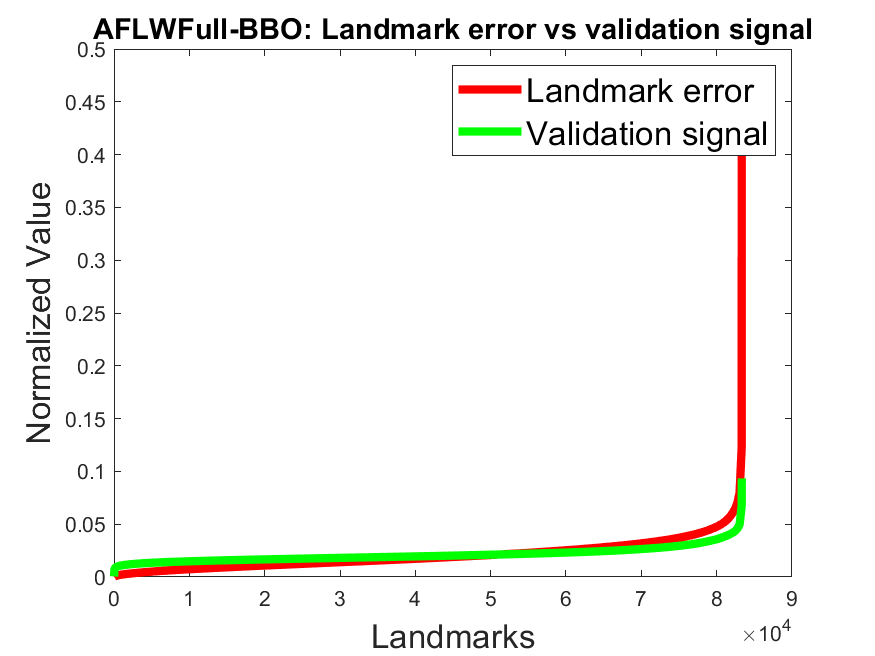}
		\includegraphics[width=0.27\linewidth]{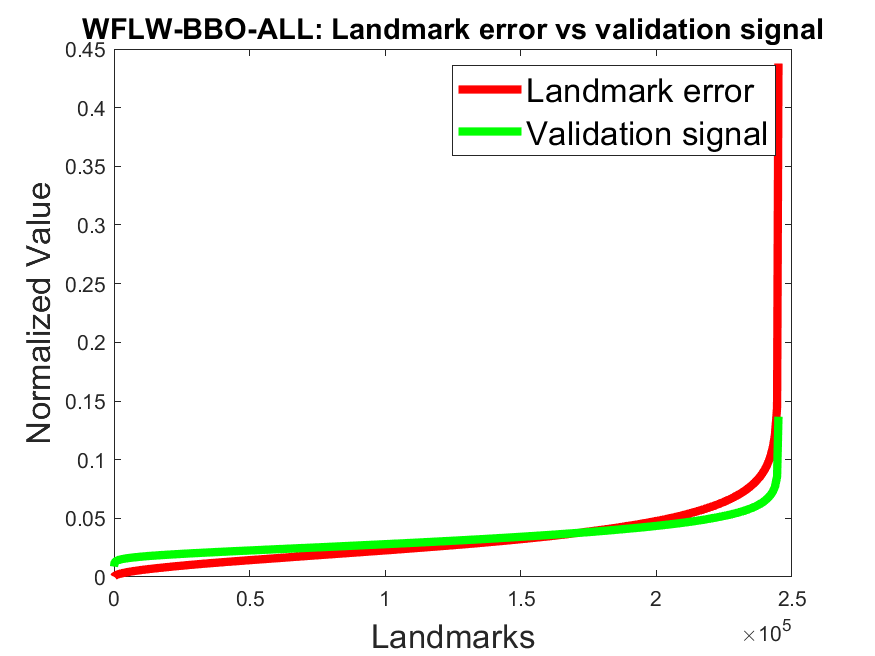}
		\includegraphics[width=0.27\linewidth]{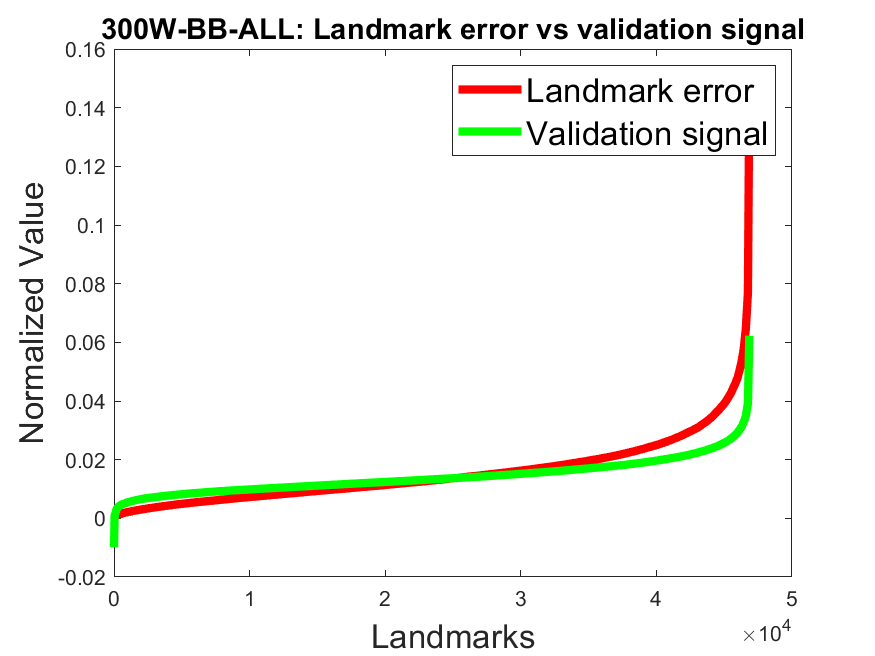}
		\includegraphics[width=0.27\linewidth]{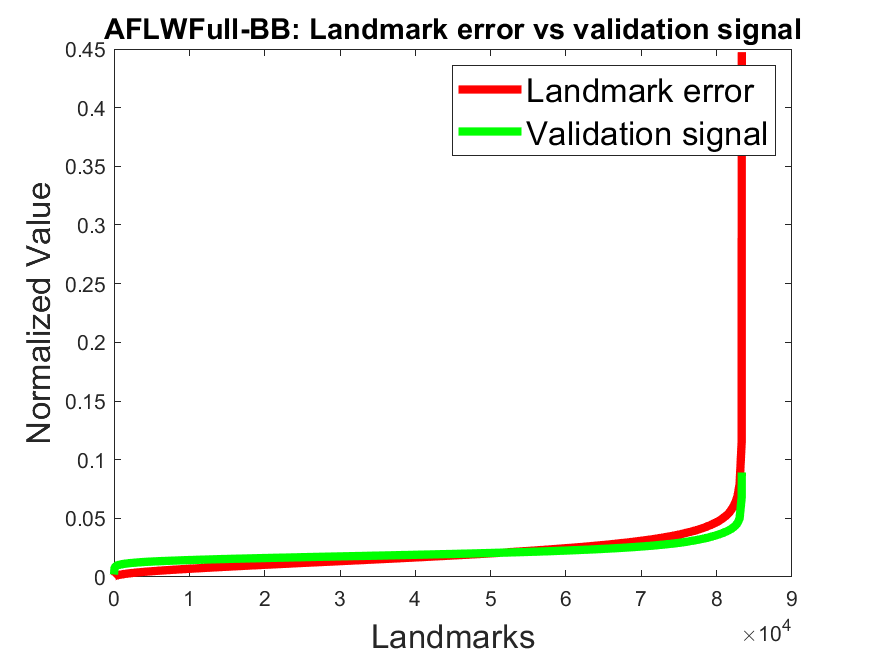}
		\includegraphics[width=0.27\linewidth]{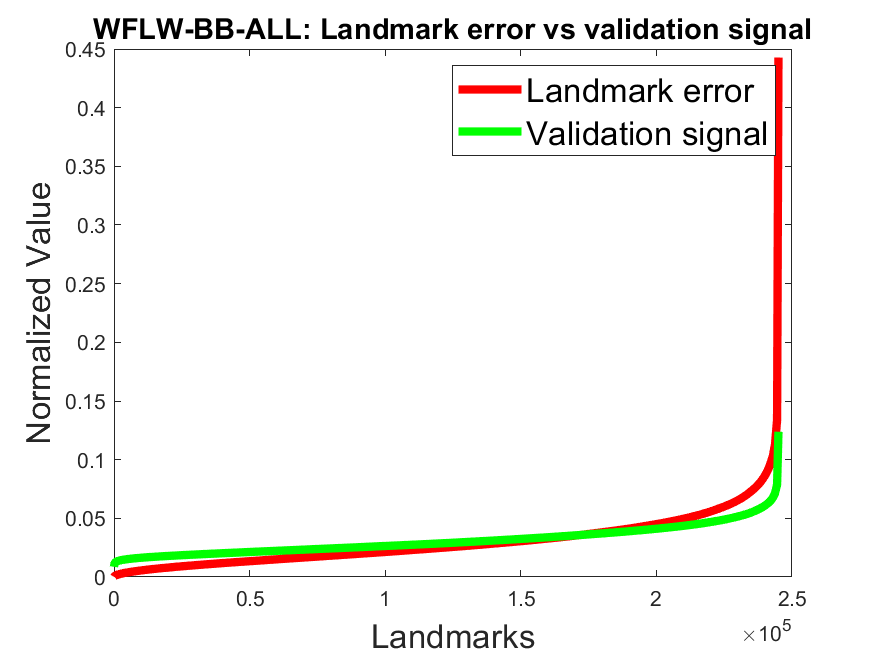}
	\end{center}
	\caption{Plots for the validation signal (green) to each landmark point error(red) in ascending order of the error(red). The first, second, and third row are the plots for the 300W, AFLW, and WFLW data set respectively. Each row shows the results for the models BBO and BB from left to right.}
	\label{fig:validacc}
\end{figure}

\textbf{Validation signal accuracy:} Figure~\ref{fig:validacc} shows the normalized error (red) against the normalized validation signal (green) for all landmarks on the datasets 300W, AFLW, and WFLW. We sorted the values ascending based on the normalized error (red) for a better visualization. For the models BBO and BB it can be seen that the validation signal correlates with the error on all datasets. To show the effectiveness of the validity signal to remove inaccurate landmarks, we made an additional experiment on the 300W and AFLW datasets. Table~\ref{tbl:cmpstoavl} shows the results for our networks by discarding different amounts of landmarks based on the validation signal. This was done by ignoring the landmarks per face with the highest validation signal values. As a baseline we also included the NON model which has no validation signal. Therefore, we selected the 10\%, 20\%, and 30\% of the landmarks per face randomly using a uniform distribution. As can be seen in Table~\ref{tbl:cmpstoavl} the model NON has only a slight improvment by discarding 30\% of the landmarks. The comparison to the models NOBB, BBO, and BB, which improve significantly, shows the efficiency of our validation loss. 

\setlength{\tabcolsep}{0.4mm}
\begin{table}
	\caption{Results in NME (smaller is better) in comparison to the state of the art on 300W and AFLW datasets together with the runtime for each model. $*$ marks models pretrained on other images or using the ground truth boundary. The best results per row are bold.}
	\begin{center}
		\begin{tabular}{l|cccccc}
			\hline
			Dataset & $*$ResNet-50~\cite{wingloss2018cvpr} & $*$LAB~\cite{wayne2018lab} & CNN-6~\cite{wingloss2018cvpr} & CNN-6/7~\cite{wingloss2018cvpr} & BB& BB \small{(Discard 30\%)}\\
			%             && (pretrained) & (pretrained) & & & \\
			%             &&  & (or ground truth boundry) & & & \\
			\hline\hline
			300W Full & 3.60 & 2.99 & 4.10 & 4.04 & 3.69 & \textbf{2.74} \\
			300W Common & 3.01 & 2.57 & 3.35 & 3.27 & 3.01 & \textbf{2.27} \\
			300W Challenging & 6.01 & 4.72 & 7.20 & 7.18 & 6.51 & \textbf{4.66} \\
			\hline
			AFLW Full & 1.47 & 1.25 & 1.83 & 1.65 & 1.56 & \textbf{1.24}\\
			\hline
			Runtime(fps) & 30 & 16,66 & \textbf{400} & 170 & 333 & 333\\
			\hline
		\end{tabular}
	\end{center}
	\label{tbl:cmplargenets}
\end{table}

\textbf{Runtime comparison:} Table~\ref{tbl:cmplargenets} shows the results on the 300W and AFLW dataset together with runtime. For our models the runtime was obtained on a NVIDIA GeForce GTX 1050ti, for LAB~\cite{wayne2018lab}, ResNet-50~\cite{wingloss2018cvpr}, CNN-6~\cite{wingloss2018cvpr} and CNN-6/7~\cite{wingloss2018cvpr} on a NVIDIA GeForce GTX Titan X. As can be seen in Table~\ref{tbl:cmplargenets} the fastest model is the CNN-6~\cite{wingloss2018cvpr} with 400fps directly followed by our architecture (333fps). The best results are optained by the model BB discarding 30\% of the landmarks per face by using the validation signal. For all landmarks the best result is obtained by LAB~\cite{wayne2018lab} with the usage of the ground truth boundary.

\section{Conclusion}
\label{sec:conclusion}
We proposed a novel loss formulation that allows regression based point estimation alongside with a reliability estimate of the result. This improves the usability of the signal as well as the accuracy of subsequent steps that rely on it. Additionally, the signal can be used for a secondary validation of face detection. To tackle the problem of unbalanced training data in available datasets, we proposed an approach to automatically balance the training data based on the produced loss of samples. The advantage of this approach is that it does not account only for the head pose but also for other challenges and imbalances in the data. 

\small
\paragraph{Acknowledgments:}
Work of the authors is supported by the Institutional Strategy of the University of T\"ubingen (Deutsche Forschungsgemeinschaft, ZUK 63). This research was supported by an IBM Shared University Research Grant including an IBM PowerAI environment. We especially thank our partners Benedikt Rombach, Martin M\"ahler and Hildegard Gerhardy from IBM for their expertise and support.

\small
\bibliographystyle{spiebib}
\bibliography{bibliography}

\end{document}